\documentclass{sig-alternate-pre}

\usepackage{algorithm}
\usepackage{algpseudocode}
\algtext*{EndWhile}
\algtext*{EndIf}
\algtext*{EndFor}
\algtext*{EndProcedure}
\usepackage{amsmath}
\usepackage{amssymb}
\usepackage{array}
\usepackage[bookmarks=false]{hyperref}
\usepackage[noadjust]{cite}
\usepackage{caption}
\usepackage{color}
\usepackage{comment}
\usepackage[inline]{enumitem}
\usepackage{epsfig}
\usepackage{etoolbox}
\usepackage{fancybox}
\usepackage{float}
\usepackage{fixltx2e}
\usepackage{graphicx}
\usepackage{hyperref}
\usepackage{listings}
\usepackage{multirow}
\usepackage{multicol}
\usepackage{placeins}
\usepackage{rotating}
\usepackage{setspace}
\usepackage[hang, tight]{subfigure}
\usepackage{threeparttable}
\usepackage{times}
\usepackage{url}
\usepackage{verbatim}
\usepackage{wrapfig}
\usepackage[usenames,dvipsnames]{xcolor}

\newbool{inccomment}
\setbool{inccomment}{true}
\newcommand{\YC}[1]{\ifbool{inccomment}{{\color{magenta}YC\@: #1}}{}}
\newcommand{\XC}[1]{\ifbool{inccomment}{{\color{blue}XC\@: #1}}{}}
\newcommand{\TD}[1]{\ifbool{inccomment}{{\color{orange}#1}}{}}
\newcommand{\FN}[1]{\ifbool{inccomment}{{\color{OliveGreen}#1}}{}}
\newcommand{\GR}[1]{\ifbool{inccomment}{{\color{Tan}#1}}{}}
\newcommand{\LD}{\ifbool{inccomment}{{\color{magenta}\\============================================\\}}}
\newcommand{\RF}{\ifbool{inccomment}{{\color{green}~[R]}}}

\newcommand{\roma}[1]{\uppercase\expandafter{\romannumeral #1\relax}}
\graphicspath{{}}

\begin{document}
\pdfpageheight 11in
\pdfpagewidth 8.5in
\title{ASP:A Fast Adversarial Attack Example Generation Framework based on Adversarial Saliency Prediction
\vspace{-6mm}
}

\author{
\alignauthor
	\large{Fuxun Yu$^1$, Qide Dong$^2$, Xiang Chen$^3$}\\
        \normalsize{The Department of Electrical and Computer Engineering\\
        George Mason University, Fairfax, VA, USA 22030\\
        \{fyu2$^1$, qdong$^2$, xchen26$^3$\}@gmu.edu}
\vspace{22mm}
}

\maketitle

\begin{abstract}
With the excellent accuracy and feasibility, the Neural Networks (NNs) have been widely applied into the novel intelligent applications and systems.
However, with the appearance of the Adversarial Attack, the NN based system performance becomes extremely vulnerable: the image classification results can be arbitrarily misled by the adversarial examples, which are crafted images with human unperceivable pixel-level perturbation. As this raised a significant system security issue, we implemented a series of investigations on the adversarial attack in this work:
	We first identify an image's pixel vulnerability to the adversarial attack based on the adversarial saliency analysis. By comparing the analyzed saliency map and the adversarial perturbation distribution, we proposed a new evaluation scheme to comprehensively assess the adversarial attack precision and efficiency.
	Then, with a novel adversarial saliency prediction method, a fast adversarial example generation framework, namely ``\textit{ASP}'', is proposed with significant attack efficiency improvement and dramatic computation cost reduction.
	Compared to the previous methods, experiments show that \textit{ASP} has at most {12}$\times$ speed-up for adversarial example generation, {2}$\times$ lower perturbation rate, and high attack success rate of {87}\% on both MNIST and Cifar10.
	\textit{ASP} can be also well utilized to support the data-hungry NN adversarial training. By reducing the attack success rate as much as 90\%, \textit{ASP} can quickly and effectively enhance the defense capability of NN based system to the adversarial attacks.
\end{abstract}

\section{Introduction}
\label{sec:intro}
Nowadays, the Neural Network (NN) is considered as one of the most representative machine learning technologies, and has been widely applied into intelligent applications and embedded systems, such as augmented reality devices \cite{AR}, mobile natural language processing \cite{NLP}, and autonomous-driving system \cite{AutoDriving}. However, a considerable security issue has also emerged recently with an NN dedicated attack method, namely, the Adversarial Attack \cite{Intriguing}.

\begin{figure}[!tb]
	\centering
	\captionsetup{justification=centering}
	\vspace{-3mm}
	\includegraphics[width=3.3in]{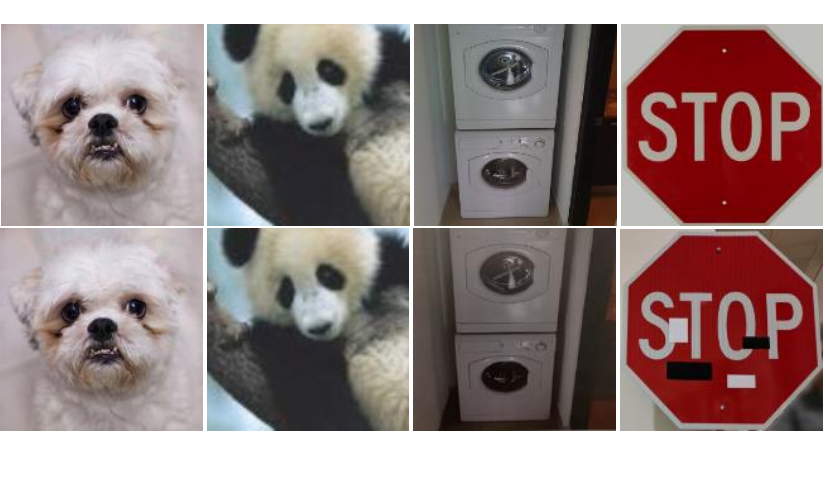}
	\vspace{-12.5mm}
	\caption{Original (Row 1) vs. Adversarial Examples (Row 2)}
	\label{fig:1}
	\vspace{-4mm}
\end{figure}

The current adversarial attacks are usually designed to manipulate the NN image classification results arbitrarily, which is achieved by injecting adversarial examples into the NN testing phase \cite{Explaining}.
Those adversarial examples are generated by dedicated adversarial attack algorithms, which distort the original images with pixel-level perturbations that even human vision can't perceive \cite{Physical}.
Even imperceptible, these perturbations can effectively fool the state-of-the-art NN systems. As shown in Fig.~\ref{fig:1}, although the human can still recognize the adversarial examples as the correct classes of the originals, the similar adversarial examples could cause $\sim$90\% misclassification rate to a well-trained NN system \cite{Intriguing}-\cite{Stopsign}.

The adversarial attacks not only demonstrate the vulnerability of the NNs with significant security issues in practical systems, but also reveal the significant cognitive difference between the NNs and the human vision, which makes the adversarial attack an important approach for the NN study. Therefore, more and more effort has been made to the adversarial attack research recently \cite{Physical}-\cite{Distillation}. However, without deep understanding and comprehensive evaluation, most of the adversarial attack methods still suffer from inconsistent attack success rate, large perturbation area, and considerable computation cost \cite{Deepfool}-\cite{Robustness}.

To gain a better understanding of the adversarial attack, we implemented a series of investigations on the adversarial attack, in terms of attack evaluation, attack generation, and attack defense. In this work, we have the following contributions:

	\vspace{1mm}

$\bullet$ We invented a comprehensive adversarial attack evaluation sche-me.
	By identifying an image's pixel vulnerability distribution to the adversarial attack with saliency analysis, we can evaluate the precision of the perturbation distribution generated on the adversarial example, and therefore the overall adversarial attack efficiency;
	\vspace{1mm}

$\bullet$ We designed \textit{ASP}, an innovative fast adversarial example generation framework.
	\textit{ASP} is based on a new adversarial saliency prediction method with comprehensive adversarial pattern analysis and extraction.
	With predicted adversarial pattern, the high-quality adversarial examples can be quickly generated without considerable computation overhead;
	\vspace{1mm}

$\bullet$ We applied \textit{ASP} to support the data-hungry adversarial training process.
	With massive generated adversarial examples included in the NN training phase as vaccines, the immunity of NNs to the adversarial attack can be effectively enhanced;
	\vspace{1mm}

$\bullet$ We implemented the proposed fast adversarial attack example generation framework, as well as the fast adversarial training framework, and quantitatively evaluated their performance comparing to the previous methods.
	\vspace{1mm}

Experiments show that, compared to the previous adversarial attack methods, \textit{ASP} has significant cost reduction with (2$\sim$100)$\times$ computation speed-up on MNIST dataset \cite{MNIST} and (2$\sim$12)$\times$ speed-up on Cifar10 \cite{cifar}. The generated adversarial examples also demonstrated optimal quality with (1.5$\sim$2)$\times$ lower perturbation rate, and high attack success rate of 87\%$\sim$92\%. When the \textit{ASP} is utilized in adversarial training, the adversarial attack can be effectively defended with dramatic attack success rate reduction of 45\%$\sim$90\%.



\section{Preliminary}
\label{sec:prelim}
\vspace{-2mm}

\begin{figure}[!tb]
	\centering
	\captionsetup{justification=centering}
	\vspace{-3mm}
	\includegraphics[width=3.3in]{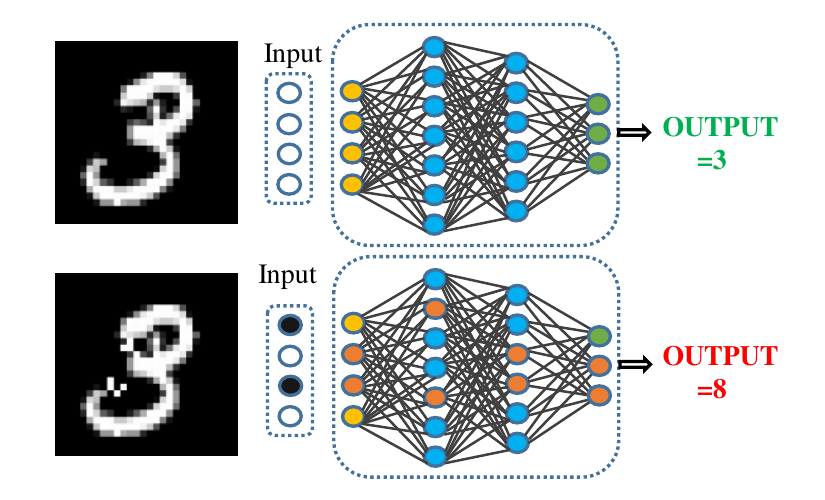}
	\vspace{-8mm}
	\caption{Gradients based Adversarial Attack}
	\label{fig:2}
	\vspace{-5mm}
\end{figure}

\subsection{Gradient-based Adversarial Attacks}
An NN could be seen as a large-scale non-linear function, composed with massive volumes of neurons, weights ($w$) and bias values ($b$). For the $k^{th}$ neuron \textit{i} in layer \textit{l}, the activation $\alpha$ is:
\begin{equation}
	\medmuskip=-1mu
	\alpha = \delta(\sum\nolimits_{k} {w_{k}^{l}\alpha_{k}^{l-1}+b_{k}^{l}}).
	\label{eq:1}
\end{equation}

While, the whole NN could be written as a function of \textit{f(x)}:
\begin{equation}
\begin{split}
	\medmuskip=-1mu
	&f(x) = \\
	&\delta(\sum\nolimits_{k} {w_{k}^{l}\delta(\sum\nolimits_{k} {w_{k}^{l-1}\delta(..\delta(\sum\nolimits_{k} {w_{k}^{1}x_{k}^{0}+b_{k}^{0}})..)+b_{k}^{l-1}})+b_{k}^{l}}).
	\label{eq:2}
\end{split}
\end{equation}

The training phase of a NN can be seen as a loss function optimization process for $f(x)$, which is to reduce the error differentiated between the predicted labels and the true ones. When the error is iteratively reduced by modifying the weights with gradient-based backpropagation \cite{BP}, the NN classification accuracy correspondingly increases and finally reaches a satisfaction degree.

On the other hand, the adversarial attack is similar to the training phase but with the opposite object.
As shown in Fig.~\ref{fig:2}, when certain attack targeted error is injected from the very back-end of the classification, the false gradients will be propagated eventually into the image and cause pixel-level perturbation. And during the testing phase of forward-propagation, the distorted image will cause corresponding classification failure, in other words, adversarial attack success.
	
To achieve the optimal image perturbation, we suppose $f: R^{m} \to {[1...y]}$ is a classifier mapping an $m$-dimensional input vector $R^{m}$ to a discrete label set. For an original image of $x \in R^{m}$ and an attack targeted false label $l \in {[1...y]}$, the adversarial example generation can be defined as a perturbation minimization process:
\begin{equation}
\begin{split}
	\medmuskip=-1mu
	Min&imize\ ||\theta||_{p} \ subject \ to: \\
	&1.\ f(x+\theta) = l \\
	&2.\ x^{adv}=x+\theta, x^{adv} \in [0,1]^{m},
	\label{eq:3}
\end{split}
\end{equation}
where, $x$ and $x^{adv}$ are the original image and adversarial example respectively, vector $\theta$ is the perturbation vector, $p$ is the regulation factor ($p=1$ for \textit{L1-norm},  $p=2$ for \textit{L2-norm}, and $p=+\infty$ for \textit{L-$\infty$-norm}), and [0, 1] is the image pixel value bound constraint. ([0,1] is normalized from $0\sim255$ for MNIST and Cifar10.)

\subsection{State-of-the-Art Attacking Methods}
\begin{figure}[!b]
	\centering
	\captionsetup{justification=centering}
	\vspace{-3mm}
	\includegraphics[width=3.3in]{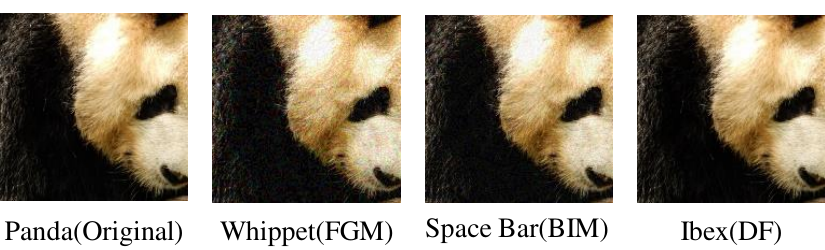}
	\vspace{-5mm}
	\caption{Adversarial Examples Predction Results}
	\label{fig:3}
	\vspace{-1mm}
\end{figure}

Currently, several representative adversarial attack methods are proposed for adversarial example generation, such as Fast Gradient Method (\textit{FGM}) \cite{Physical}, Basic Iterative Method (\textit{BIM}) \cite{Physical}, DeepFool method \cite{Deepfool}, and etc. All these methods can generate effective adversarial examples with human unperceivable pixel-level perturbation as shown in Fig.~\ref{fig:3}.

\textit{Fast Gradients Method}:
\textit{FGM} is one of the simplest and fastest adversarial attack methods. It utilizes a relatively big perturbation step with a hyper-parameter of $\epsilon$ to generate perturbation without specifically targeted false label:
\begin{equation}
	\medmuskip=-1mu
	x^{adv}\ =\ x\ +\ \epsilon sign\bigtriangledown_{x} J(x,y_{true}),
	\label{eq:4}
\end{equation}
where, $J()$ is the NN loss function in terms of cross-entropy usually. As a non-targeted adversarial attack, such a method only costs one call to back propagation, offering high attacking speed \cite{Physical}.

\textit{Basic Iterative Method}:
\textit{BIM} is an improved method composed of iterative \textit{FGM}, which runs \textit{FGM} multiple times with a small step size $\alpha$. During iteration pixel values need to be clipped to [0, 1] after each iteration to ensure that they are in an $\epsilon$-neighborhood of the original image \cite{Physical}.
\begin{equation}
	\medmuskip=-1mu
	x_{N+1}^{adv}\ =\ Clip_{x,\epsilon} \{{x_{N}^{adv}\ +\ \alpha sign\bigtriangledown_{X} J(x_{N}^{adv},y_{true})}\}
	\label{eq:5}
\end{equation}
This method is a non-targeted attack but takes more iterations compared to \textit{FGM}. Both \textit{FGM} and \textit{BIM} are optimized for \textit{L-$\infty$-norm} because they need to restrict the perturbation step to be smaller than a certain threshold $\epsilon$.

\textit{DeepFool}:
\textit{DeepFool} is also a non-targeted attack technique optimized for the \textit{L2-norm}. This method supposes that a NN is a linear functions with a hyperplane separating each class from another. By this assumption, it analytically derives the optimal perturbation to push the example to pass the hyperplane. Since NNs are not actually linear, it repeats this process iteratively until a successful adversarial example is found.

\subsection{Adversarial Training for Defense}
For one certain adversarial example, the adversarial attack succeeds when the NN wrongly recognizes the hidden adversarial pattern to be the main pattern. These circumstances happen because NNs cannot generalize well on pictures containing both adversarial and main patterns. Thus, training on both adversarial and original examples could be seen as one of the data augmentation schemes: augmentating the training data to improve the generalization of NNs to the adversarial examples. Previous works show that, when adding adversarial examples as a ``vaccine'' subset in the training data, the NN could be more immune to the adversarial attacks with significantly lower attack success rate \cite{Explaining}. In this work, we also applied the adversarial training to enhance the NN immunity to the adversarial attack, more details will be presented in Section~\ref{sec:train}.


\section{Performance Evaluation for \\ Adversarial Attacks}
\label{sec:eve}

In this section, we will investigate the performance of current adversarial attacks, and propose a novel comprehensive evaluation scheme based on the adversarial attack analysis.

\subsection{Current Evaluation Metrics}
In the current adversarial attack works, the evaluation metrics mainly focus on the error manipulation levels, such as: \textit{Attack Success Rate} (1 - \textit{Prediction Accuracy}), \textit{Perturbation Rate} (ratio of the manipulated pixel number to the image resolution), and \textit{Perturbation Degree} (ratio of the total manipulated pixel value to the overall image pixel value summation) \cite{Deepfool}-\cite{Robustness}. Table 1 shows how these metrics evaluate the performance of \textit{FGM}, \textit{BIM} and \textit{DeepFool} attacking the MNIST dataset \cite{MNIST}.
\begin{table}[t]
	\centering
	\caption{Evaluation with Current Metrics}
	\vspace{-2mm}
	\small{\begin{tabular}{|c|c|c|c|} \hline
		&\textit{FGM}&\textit{BIM}&\textit{DeepFool}\\ \hline
		Attack Success Rate&92.4\%&99.0\%&98.9\%\\ \hline
		Perturbation Rate&71.85\%&63.78\%&65.92\%\\ \hline
		Perturbation Degree&20.63\%&13.64\%&3.41\%\\ \hline
	\end{tabular}}
\vspace{-5mm}
\label{tab:1}
\end{table}

However, those metrics can't fully explain the effectiveness of those attacking methods. For example, \textit{FGM} has both the highest perturbation rate and perturbation degree, which should lead a highest attack success rate, since more perturbation means a higher chance of successful attack. But, in fact, \textit{FGM} has the lowest success rate of 92.4\%, compared to \textit{BIM} of 99.0\% and \textit{DeepFool} of 98.9\%. On the other hand, \textit{DeepFool} has much lower perturbation degree of 3.41\% and medium perturbation rate of 65.92\%, but achieves the best success rate of 98.9\% as \textit{BIM}. Moreover, even \textit{BIM} and \textit{DeepFool} have the similar success rate, \textit{BIM} has 4$\times$ more perturbation degree than \textit{DeepFool}.

\subsection{Adversarial Saliency Efficiency}
In this work, we propose a new evaluating scheme -- \textit{Adversarial Saliency Efficiency (ASE)}. Rather than only analyzing the attack result, our \textit{ASE} examines adversarial attack with precision and efficiency with adversarial saliency analysis \cite{Limitation}.
The intuitive of this scheme is to tell if the adversarial example generation algorithm can precisely find the most vulnerable or sensitive pixels to cast perturbation. Here, the sensitivity is defined as how much prediction results error that a unit perturbation in the pixel could cause.

Mathematically, the \textit{ASE} is derived from the pixel saliency analysis based on \textit{Jacobian-matrix}, which describes the pixel vulnerability distribution for classification $f(x_{n}): {[y_0,y_1...y_m]}$ \cite{Limitation}:
\begin{equation}
\begin{split}
	\medmuskip=-1mu
	&\frac{\delta y_0}{\delta x_0},\ \frac{\delta y_0}{\delta x_1}\  \ \cdots \ \ \ \frac{\delta y_0}{\delta x_n} \\
	J_{f(x)}(x_0,x_1 ... x_n) = [ &\ \ \ \vdots\ \ \ \ \ \ \ \  \vdots\ \ \ \ \  \dots\ \ \ \ \  \frac{\delta y_1}{\delta x_n}  ] , \\
	&\frac{\delta y_m}{\delta x_0},\ \frac{\delta y_m}{\delta x_1}\  \cdots \ \ \frac{\delta y_m}{\delta x_n}
	\label{eq:6}
\end{split}
\end{equation}
where, $\frac{\delta y_m} {\delta x_n}$ means the pixel's differential impact for the image to be classified as label $y_m$.
Based on the \textit{Jacobian-matrix}, \textit{Adversarial Saliency Map(ASM)} is defined by the following equation:
\begin{equation}
\begin{split}
	\medmuskip=-1mu
	SaliencyMaps = mask * (\frac{\delta y_{true}} {\delta x_n}) * (\sum{\frac{\delta y_{false}} {\delta x_n}})
	\\
	where\ mask=\left\{
			\begin{aligned}
			1 &, if \frac{\delta y_{true}} {\delta x_n} < 0 \ and\ \sum{\frac{\delta y_{false}} {\delta x_n}} > 0\\
			0 &, if \frac{\delta y_{true}} {\delta x_n} \ge 0 \ or\ \sum{\frac{\delta y_{false}} {\delta x_n}} \le 0
			\end{aligned} ,
			\right.
	\label{eq:7}
\end{split}
\end{equation}
where, a mask scheme is proposed to polarize the robust pixels and attack robust pixels (with true label's derivative score $\le0$ or the sum of false label's derivative score $\ge0$). Hence, the \textit{ASM} can offer a comprehensive statistic of the vulnerable pixels of an image.

With \textit{ASM}, \textit{ASE} is then calculated by the divergence of \textit{ASM} distribution and adversarial perturbation distribution generated by specific adversarial attack method, which is defined as:
\begin{equation}
\begin{split}
	\medmuskip=-1mu
	ASE = \frac{\sum_i^N{(Sort(SaliencyMap)[i]*Perturbation[i])}} {\sum_j{|Perturbation[j]|}}.
	\label{eq:8}
\end{split}
\end{equation}

In Eq.~\ref{eq:8}, rather than evaluate every individual pixel in the adversarial attack, we choose the $N$ most vulnerable pixels with the highest derivative score. The $N$ is selected in regards of the resolution of the input image, since an improper value could cause significant interpretability issue of \textit{ASE}. In our experiment, we choose $N$=50 for the MNIST dataset and $N$=100 for the Cifar10 dataset. The sum of effective perturbations on the $N$ pixels will be further normalized by $\sum_j{|Perturbation[j]|}$, which is the perturbation summation of all the pixels on the image regardless of their vulnerability.

\begin{figure}[!tb]
	\centering
	\captionsetup{justification=centering}
	\vspace{-3mm}
	\includegraphics[width=3.3in]{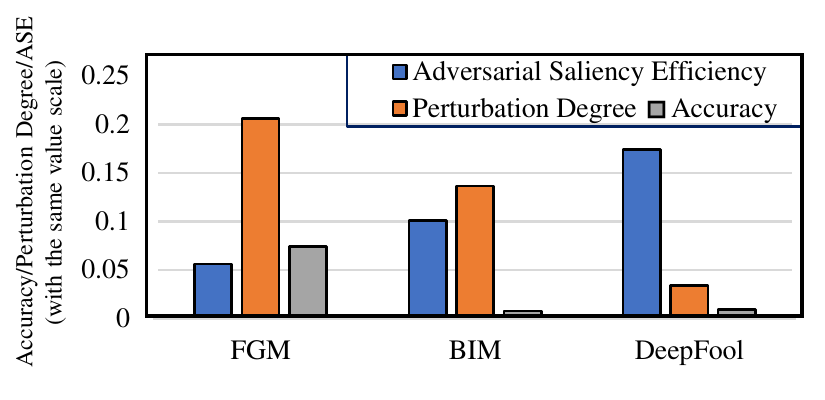}
	\vspace{-8mm}
	\caption{Adversarial Saliency Efficiency Evaluation}
	\label{fig:4}
	\vspace{-5mm}
\end{figure}

To evaluate the effective of the proposed \textit{ASE}, we applied it to three methods as shown in Fig.~\ref{fig:4}. The \textit{ASE} of the three methods are 5.6\%, 10.1\%, 17.9\% respectively. And we can see that the \textit{ASE} describes their efficiency very well:
For \textit{FGM}, even its perturbation rate is very high, the ASE is low, which means most of the perturbation are not useful for attacking purpose. This causes the classification accuracy higher than other two attacks, which means lower attack success rate. On the other hand, the low perturbation rate and high \textit{ASE} of \textit{DeepFool} indicate an outstanding pixel attack precision, which make the attack success rate optimal. For \textit{BIM}, its \textit{ASE} along with its perturbation degree are both relatively high, indicating the best performance with precise and concentrated attack.

With our proposed evaluation scheme, the adversarial attack efficiency can be comprehensively evaluated and applied to guide the fast adversarial attack as well as defense.

\section{ASP Framework Design}
\label{sec:algo}
Due to the complex pixel vulnerability analysis and backpropagation computation, the current adversarial attack methods all suffer from high computation cost. In this work, we propose our fast adversarial example generation framework \textit{ASP}: With previous \textit{ASE} analysis, we first propose a new adversarial saliency prediction method, which can effectively analyze and extract a general adversarial pattern for dedicated attack. With the prediction, the dedicated pixel analysis for each adversarial example generation can be effectively avoided and significantly improve the attack speed.

\subsection {Adversarial Saliency Prediction}
From previous analysis, we have known that \textit{ASM} score could be used to find the most sensitive or vulnerable pixels to attack. And most adversarial attack should follow the specific adversarial pattern along with those vulnerable pixels distribution.
	An example is shown in Fig.~\ref{fig:2}, which is an adversarial example attacking MNIST classificatoin with a target of 3 $\rightarrow$ 8. The best attack performance is achieved by perturbing the most vulnerable pixels mainly lying in the different trace of 3 $vs. $8. In fact, not only for 3 $\rightarrow$ 8, for any other certain pair of original class and target class, their main patterns should be analogous.

Therefore, a general pattern in \textit{ASM} should be predictable for one certain pair of classes, which could be utilized for attack analysis and computation optimization.
	Suppose we have a N class dataset. For each pair of $l_{ori} \rightarrow l_{target} (l_{ori} \ne l_{target})$, we could predict the \textit{ASM} pattern of them by producing large numbers of \textit{ASM} training dataset, based on which we could use linear regression algorithm to predict the \textit{ASM} distribution in the general pattern of saliency maps of each pair. The detailed algorithm for the adversarial saliency prediction is shown in Algorithm 1.

\begin{figure}[t]
	\centering
	\captionsetup{justification=centering}
	\vspace{-3mm}
	\includegraphics[width=3.3in]{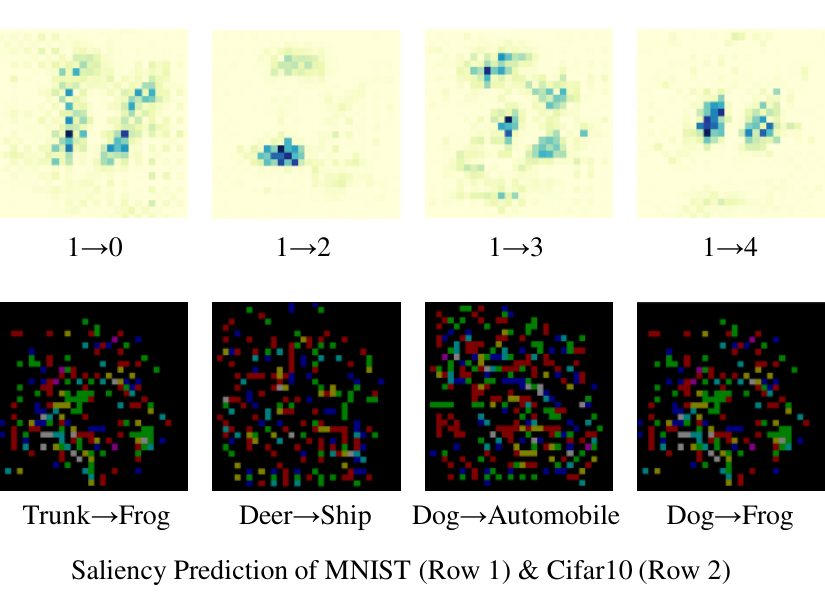}
	\vspace{-7mm}
	\caption{Saliency Prediction Examples}
	\label{fig:5}
	\vspace{-5mm}
\end{figure}

A sample prediction for MNIST and Cifar10 is shown in Fig.~\ref{fig:5}. For MNIST dataset, we could see clear patterns for a series of target adversarial saliency, which are quite easy to understand: Attacking these pixels will alter the original handwriting shape towards the target class's shape. For Cifar10 dataset, since the images have three channels (RGB), the pattern is not that straight-forward to understand but these patterns all perform well in the following attack phase, which will be discussed in Section~\ref{sec:eval}.

\subsection{Fast Adversarial Example Generation}
\begin{figure}[b]
	\centering
	\captionsetup{justification=centering}
	\vspace{-3mm}
	\includegraphics[width=3.3in]{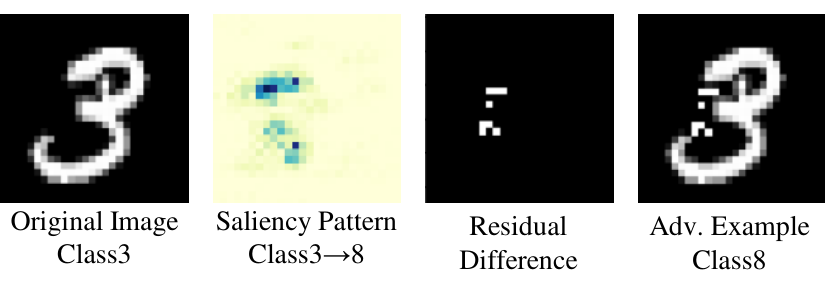}
	\vspace{-7mm}
	\caption{Process of Fast Adversarial Example Generation}
	\label{fig:6}
	\vspace{-3mm}
\end{figure}

With predicted adversarial pattern, attackers can directly utilize it to replace the time-consuming gradients computation for individual adversarial example generation. Specifically, we first distort the most sensitive pixels value according to the order of our ASP pattern score and a certain perturbation rate.
	Most examples will succeed in causing misclassfication results after this step. Fig.~\ref{fig:6} shows such one adversarial image generation example in MNIST. From it we could see that the predicted pattern is still quite accurate and matches the intuitive adversarial pattern. In addition, for unsuccessful attacked images, we will further adaptively distort more pixels. The overall algorithm is shown in Algorithm 1.

To meet the highest performance of adversarial attack effectiveness, we further optimize the adversarial example generation process. We know that a higher perturbation rate not only  increases the attack success rate but also makes the adversarial feature perceivable.
	So, to analysis the trade-off between the perturbation rate and attacking success rate, we made a series of tests and eventually choose 21.7\% as the perturbation rate, resulting in 1.4\% of test accuracy (or 98.6\% of attacking success rate) with only a few unsuccessful examples.

For those ineffective examples, we further adaptively increase their perturbation rate and modifying pixels according to \textit{ASM} scores by a step of 10 in prediction test iteration. Once the prediction results change to false, we stop the process.  Since number of these examples is minimum, the adaptive perturbation process would not take considerable time or influence the average perturbation rate to a large extent. Thus, we could make sure that our attacking method could achieve as high attacking successful rate and maintain low perturbation rate as possible.

\subsection{ASP Framework Overview}
In this section, we combine the previously proposed algorithms and methods to form an effective fast adversarial example generation framework, namly \textit{ASP}:

During the prediction training: we first use chain forward derivative to calculate the \textit{Jacobian-matrix} for each $x$ with different false label $y$ in training set.
	Then we build the saliency maps based on the \textit{Jacobian-matrix}. With the saliency maps training data, we use high performance server to predict the map pattern for each pair $x \rightarrow y$. We call the above steps the training session.

During the adversarial attack: with the perturbation distribution on the predicted general pattern, we could directly apply perturbation on the image under attack.
	This will cost much less resources than other algorithms because of no gradients calculation. By choosing the best-fit parameters to trade-off the adversarial effectiveness and perturbation rate, the number of ineffective adversarial examples could be minimized. And additional perturbation degree will be continuously generated to the ineffective examples until successful attack.
	The overview of the proposed fast adversarial example generation framework is shown in Algorithm 1.
\begin{algorithm}[!tb]
	\caption{ASP Framework}
	\begin{algorithmic}[1]
		 \Procedure{$Prediction\ Training\ $}{$SaliencyPattern$}
		 	\State $i=0, label(X) = x;$
			\For {${X} \in {TrainingSet\ S}$}
				\vspace{0.5mm}
				\For {${y} \in {TrainingLabel\ L}\ and\ y \ne x$} 
					\State $Maps[x][y][i] \gets ASM(F(X),y);$
					\State i++;
				\EndFor
			\EndFor
			\For {${(x,y)} \in {L \times L}$}
				\If {$y != x$}
					\State $SaliencyPattern[x][y] \gets LR(Maps[x][y]);$
				\EndIf
			\EndFor
		\EndProcedure
		\Return $SaliencyPattern$
		\Procedure{$Generating \ Adversarial\ Examples$}{}
			\For {${X} \in {TestSet\ S'}$}
				\For {${y} \in {TestLabel\ L'\ } and {\ y\ \ne\ x}$}
					\State $List = Sort(SaliencyPattern[x][y])$
					\For {${i} \in {List[0:M]}$} \Comment{Perturbating List}
						\State $X_{adv}[i] = Clip_{(0,1)}(X[i] + \epsilon);$
					\EndFor
				\EndFor
				\While {$f(x_{adv}) = x\ and\ List[M] != NULL$}
					\For {${i} \in {List[M:M+10]}$}\Comment{If Attack Failed}
						\State $X_{adv}[i] = Clip_{(0,1)}(X_{adv}[i] + \epsilon);$
					\EndFor
				\EndWhile
			\EndFor
		 \EndProcedure
		 \Return $x_{adv}$
	\end{algorithmic}
	\label{alg:1}
\end{algorithm}

\section{ASP based Adversarial Training}
\label{sec:train}

As aforementioned, the adversarial training can effectively enhance the NN defense capability. In this section, we propose an \textit{ASP} fast adversarial training framework with the proposed \textit{ASP}.

\subsection {Mechanism of Adversarial Training}
\vspace{-0.5mm}
The adversarial training can be seen as a network regulation process with data augmentation scheme, which augments the training data with adversarial examples to improve the NN generalization capability for better tolerance with the adversarial examples:

The loss functions of a normal training and an adversarial training can be formulated as:
\begin{equation}
	\medmuskip=-1mu
	Loss=J(\sigma, x, y);  Loss_{adv} = J(\sigma, x+\triangle_x, y),
	\label{eq:9}
\end{equation}
where, $\sigma$ in the parameter set of the NN, $\triangle_x$  is the perturbation value from the adversarial attack, and therefore $x+\triangle_x$ represents the adversarial example.
	By integrating these two loss functions, the adversarial training procedure could be formulated as minimizing the following function:
\begin{equation}
	\medmuskip=-1mu
	J_{adv}(\sigma',x,y) = \alpha \times J(\sigma', x, y) + (1-\alpha) \times J(\sigma', x+\triangle_x, y),
	\label{eq:10}
\end{equation}
where, $\alpha$ is the parameter set to adjust the weight ratio between original dataset and adversarial examples, for which we choose $\alpha=0.5$, considering that both original dataset and adversarial examples are equally important.

This adversarial training process will eventually minimize the prediction error as well as the adversarial attack success rate. However, from Eq.~\ref{eq:10}, we can also tell that, the adversarial training is also a data-hungry process for the adversarial examples. Considering the low adversarial example generation speed with current adversarial attack methods, the adversarial training efficiency is highly compromised. (The computation time for the adversarial attacks will be quantitatively investigated in Section~\ref{sec:eval}.)

\subsection {Fast Adversarial Training Framework}
\begin{figure}[!t]
	\centering
	\captionsetup{justification=centering}
	\vspace{-2mm}
	\includegraphics[width=2.8in]{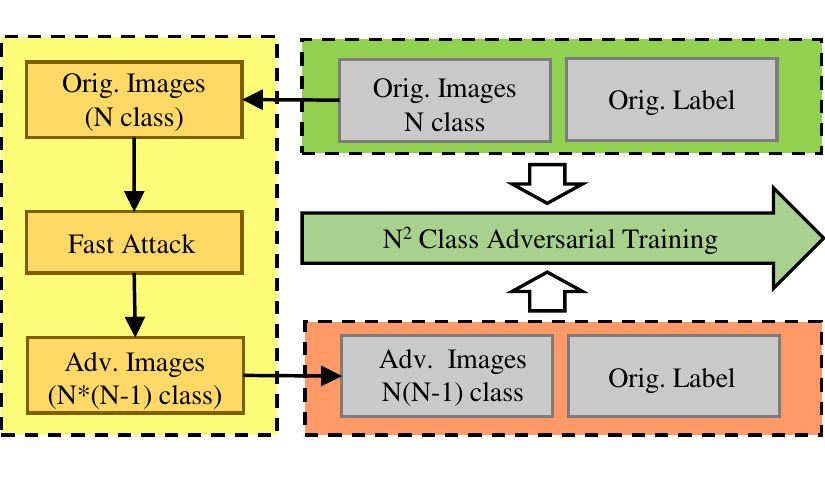}
	\vspace{-5mm}
	\caption{Adversarial Training Process}
	\label{fig:7}
	\vspace{-5mm}
\end{figure}

The proposed high performance \textit{ASP} provides an optimal solution to the data-hungry adversarial training process.
As shown in Fig.~\ref{fig:7}, we combine the fast adversarial example generation with adversarial training to improve the defense capability of the NN.

Suppose we have an N class dataset. We first utilize our fast adversarial example generation method \textit{ASP} to attack on the training dataset samples. Specifically, \textit{ASP} method will produce $N-1$ adversarial examples for each training sample. With our \textit{ASP} method, large amounts of adversarial examples could be produced efficiently. Combining these adversarial examples with original label, we could get an adversarial training dataset with $N$ times size of the original dataset (N class for original and N(N-1) class for adversarial examples). Then adversarial loss function is used to train the neural network. When the adversarial training phase is done, the neural network will become more robust to adversarial attack, which will be discussed in Evaluation part.

\section{Performance and Evaluation}
\label{sec:eval}
	\vspace{-2mm}

\subsection {Experiment Setup}
We mainly test our algorithms on two most popular dataset for image classfication -- MNIST and Cifar10. For MNIST hand writing digits classification, we use a four-layer convolution neural network as test object, which contains 3 convolutional layer with ReLu activation function and 1 fully-connected layer with \textit{SoftMax} function as output. This model could achieve 99.2\% classification accuracy after training 10 epochs. For Cifar10 image classification, we use a ten-layer convolutional neural network training with dropout technique as test object, which includes 6 convolutional layers, 3 max pooling layers and 1 fully-connected layer with \textit{SoftMax} function as output. This model achieves 85\% percents classification accuracy on test images after training 100 epochs. We evaluate the performance of different attack methods all based on these two models within the same test environment: Tensorflow-1.3 \cite{Tensorflow} with CUDA support, GTX1080 8G. In addition, for computation time evaluation, we set the parameters in all algorithms to cause just above 90\% and 85\% misclassification rates for MNIST and Cifar10 to ensure they produce the same adversarial effect. The adversarial attacks \textit{FGM}, \textit{BIM} and \textit{DeepFool} are tested by using v2.0.0 of CleverHans library\cite{cleverhans}.

\subsection {Performance of ASP on MNIST}
\begin{figure}[!t]
	\centering
	\captionsetup{justification=centering}
	\vspace{-3mm}
	\includegraphics[width=3.3in]{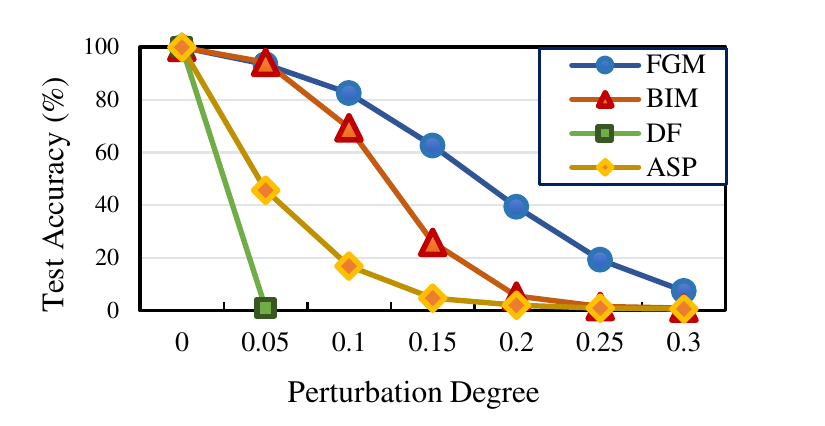}
	\vspace{-7mm}
	\caption{Test Accuracy Comparation on MNIST}
	\label{fig:8}
	\vspace{-4mm}
\end{figure}

In this section, we compare our \textit{ASP} algorithm with current existed algorithms: \textit{FGM}, \textit{BIM} and \textit{DeepFool}, with evaluation metrics of perturbation rate, attack success rate, computing cost and \textit{ASE.} First, regarding different perturbation degree, the results are shown in {Fig.~\ref{fig:8}}.
As {Fig.~\ref{fig:8}} shows, under the low perturbation degree, \textit{ASP} algorithm has a better performance than \textit{FGM} and \textit{BIM}, which proves that our \textit{ASP} is more precise and effective than gradients propagation. \textit{DeepFool} algorithm achieves the best attacking performance with the lowest perturbation degree 0.05, but this comes with a much higher computation overhead. Note that \textit{DeepFool} algorithm is a heuristic searching algorithm, thus the perturbation degree is a fixed value.
\begin{figure}[!b]
	\centering
	\captionsetup{justification=centering}
	\vspace{-4mm}
	\includegraphics[width=3.3in]{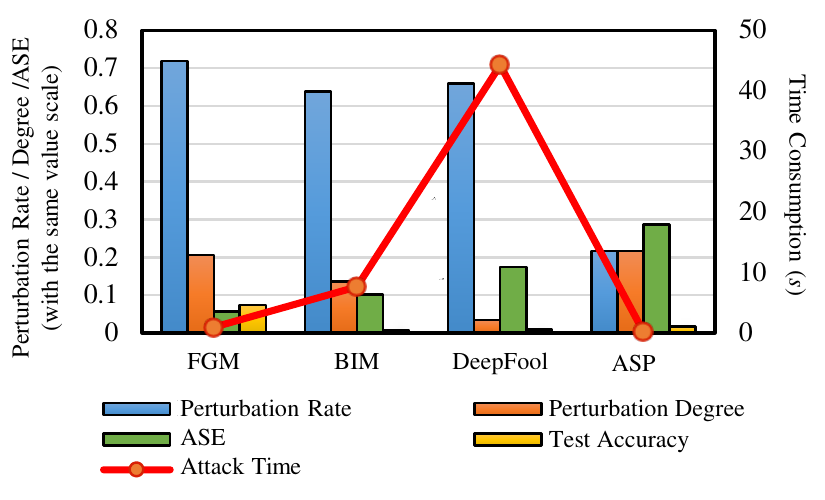}
	\vspace{-7mm}
	\caption{Performance Comparation on MNIST}
	\label{fig:9}
	\vspace{-3mm}
\end{figure}

From {Fig.~\ref{fig:9}}, first, we could clearly see that except for \textit{FGM}, all of the \textit{BIM}, \textit{DeepFool} and \textit{ASP} achieve 1\% test accuracy, which means 99\% attacking success rate. Suffering from imprecise adversarial pattern calculation, \textit{FGM}'s attacking success rate is lowest since its test accuracy is highest of all, 7.2\%. Another significant result is that \textit{DeepFool} needs much more computation time for attacking 1000 MNIST images compared to other algorithms. In fact, \textit{DeepFool} uses 44.3s for attacking 1000 MNIST images while our \textit{ASP} algorithm only uses 0.44s. By comparison, \textit{FGM} and \textit{BIM} use 0.94s and 7.65s respectively. Benefitting from the pre-trained pattern, \textit{ASP} algorithm has the shortest attacking time and also lowest computation requirements. In addition, \textit{ASP} algorithm also has the lowest perturbation rate (3 times less than all other algorithms) and highest \textit{ASE}, which also proves ASP prediction's effectiveness.

\subsection {Performance of ASP on Cifar10}
\begin{figure}[!t]
	\centering
	\vspace{1mm}
	\captionsetup{justification=centering}
	\includegraphics[width=3in]{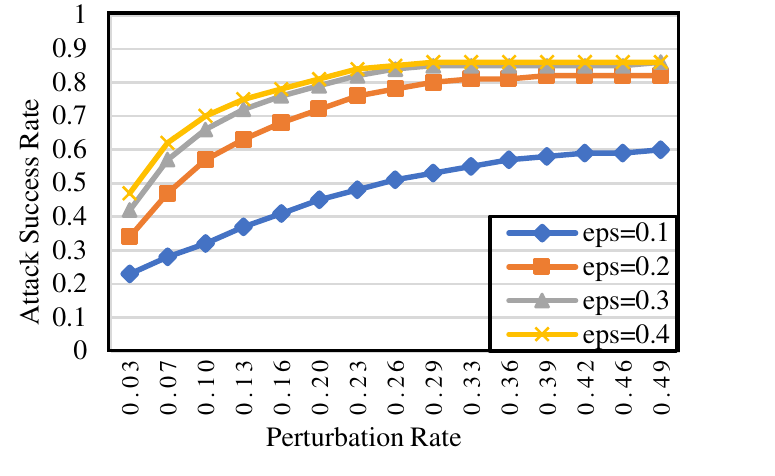}
	\vspace{-3mm}
	\caption{ASP Performance on Cifar10}
	\label{fig:10}
	\vspace{-3mm}
\end{figure}

For Cifar10 dataset, we first evaluate our \textit{ASP} attack success rate under different perturbation rates, shown in {Fig.~\ref{fig:10}}. Here $eps$ means the perturbation step size for the pixels. As {Fig.~\ref{fig:10}} shows, with just 20\% to 30\% perturbation rate and 0.2 perturbation step, ASP could achieve 85\% attack success rate. This indicates that the predicted saliency pattern in \textit{ASP} framework is also able to implement accurate and effective attack on more complex images.

\begin{figure}[t]
	\centering
	\captionsetup{justification=centering}
	\includegraphics[width=3.1in]{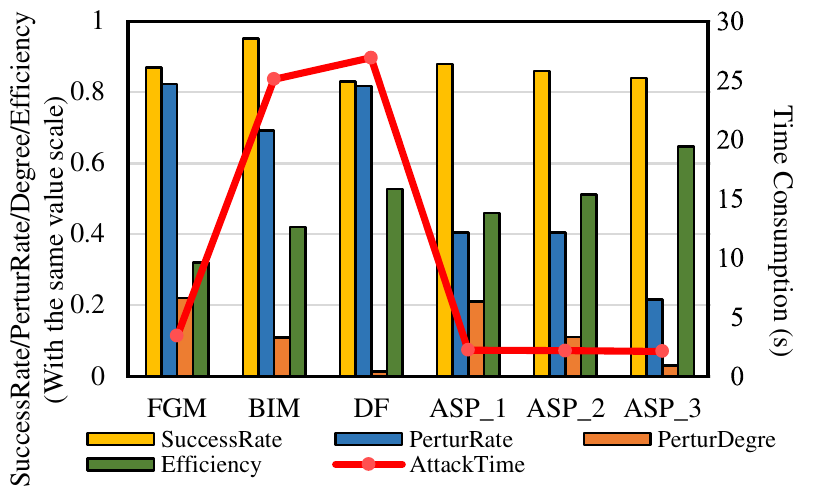}
	\vspace{-4mm}
	\caption{Performance Comparison on Cifar10}
	\vspace{-1.5mm}
	\label{fig:11}
\end{figure}

For performance comparation, we use the same perturbation step size $eps=0.3$ for \textit{FGM} and \textit{BIM}. But due to the different attack mechanisms, different perturbation degree is produced. Thus in order to compare their performance more precisely, we use three different paramter sets for \textit{ASP} so that they produce nearly same perturbation degree with our benchmark algorithms. Specifically, ASP\_1 produces the same perturbation degree (21\%) with \textit{FGM}, and for ASP\_2 and \textit{BIM} (11\%), ASP\_3 and \textit{DeepFool}(3\%). {Fig.~\ref{fig:11}} shows the \textit{ASP} performance compararation results. For all three sets, \textit{ASP} achieves same or better attack success rate (all over 83\%). In addition, perturbation rates of all three \textit{ASPs} are $2\times$ less than their counter parts, which means \textit{ASP} attack is generated more precisely and concentrated. Most importantly, \textit{ASP} takes only 2.1s to attack 1000 images, which is $1.5\times$ faster than \textit{FGM} (3.4s) and $12 \times$ faster than \textit{BIM} (25.1s) and \textit{DeepFool} (26.9s).
\begin{table}[b]
	\centering
	\vspace{-3mm}
	\caption{Adversarial Training Performance}
	\vspace{-2mm}
	\small{\begin{tabular}{|c|c|c|c|} \hline
		Test Accuracy&\textit{\textit{FGM}}&\textit{BIM}&\textit{ASP}\\ \hline
		Normal Test on original DNN&99.0\%&99.0\%&99.0\%\\ \hline
		Ad.Test on original DNN&7.38\%&0.79\%&1.71\%\\ \hline
		Ad.Test on Ad-trained DNN&96.5\%&52.3\%&45.94\%\\ \hline
	\end{tabular}}
\label{tab:2}
\end{table}

In summary, \textit{ASP} significantly outperforms \textit{FGM}, \textit{BIM} and \textit{DeepFool} on both MNIST and Cifar10 with (2$\sim$3)$\times$ lower perturbation rates, 1.5$\times$ higher attack efficiency, and most importantly, (12$\sim$100)$\times$ attack speed-up at most.

\subsection{Adversarial Training Performance}
We utilized \textit{FGM}, \textit{BIM} and \textit{ASP} algorithms and generated large amounts of adversarial examples on MNIST dataset for training purpose. Together with adversarial training object function, we re-trained our previous model with the augemented training dataset. Table 2 shows the accuracy of a NN before and after adversarial training, which indicates adversarial training could effectively enhance the defense capability of NN to the adversarial attacks.

\section{Conclusion}
\label{sec:conc}
In this work, we proposed a new fast adversarial example generation framework based on adversarial saliency prediction. Compared with current state-of-art methods, \textit{ASP} could achieve at most $\times${12} speed-up for adversarial example generation, $\times${2} lower perturbation rate, and high attack success rate of {87}\% on both MNIST and Cifar10. In addition, we also utilized \textit{ASP} to support the data-hungry NN adversarial training, which effectively enhance the robustness of NN to the adversarial attacks by reducing the attack success rate by $45\%\sim90\%$.
\vspace{2mm}

{
\scriptsize
\bibliographystyle{IEEEtran}
\bibliography{outline}

\begin{thebibliography}{10}
\providecommand{\url}[1]{#1}
\csname url@samestyle\endcsname
\providecommand{\newblock}{\relax}
\providecommand{\bibinfo}[2]{#2}
\providecommand{\BIBentrySTDinterwordspacing}{\spaceskip=0pt\relax}
\providecommand{\BIBentryALTinterwordstretchfactor}{4}
\providecommand{\BIBentryALTinterwordspacing}{\spaceskip=\fontdimen2\font plus
\BIBentryALTinterwordstretchfactor\fontdimen3\font minus
  \fontdimen4\font\relax}
\providecommand{\BIBforeignlanguage}[2]{{%
\expandafter\ifx\csname l@#1\endcsname\relax
\typeout{** WARNING: IEEEtran.bst: No hyphenation pattern has been}%
\typeout{** loaded for the language `#1'. Using the pattern for}%
\typeout{** the default language instead.}%
\else
\language=\csname l@#1\endcsname
\fi
#2}}
\providecommand{\BIBdecl}{\relax}
\BIBdecl

\bibitem{AR}
R.~T. Azuma, ``A survey of augmented reality,'' \emph{Presence: Teleoperators
  and virtual environments}, vol.~6, no.~4, pp. 355--385, 1997.

\bibitem{NLP}
J.~Hirschberg and et~al., ``{Advances in Natural language processing},''
  \emph{Science}, vol. 349, no. 6245, pp. 261--266, 1992.

\bibitem{AutoDriving}
M.~Bojarski and et~al., ``End to end learning for self-driving cars,''
  \emph{arXiv:1604.07316}, 2016.

\bibitem{Intriguing}
C.~Szegedy and et~al., ``Intriguing properties of neural networks,''
  \emph{arXiv:1312.6199}, 2013.

\bibitem{Explaining}
I.~Goodfellow and et~al., ``Explaining and harnessing adversarial examples,''
  \emph{arXiv:1412.6572}, 2014.

\bibitem{Physical}
A.~Kurakin and et~al., ``Adversarial examples in the physical world,''
  \emph{arXiv:1607.02533}, 2016.

\bibitem{Stopsign}
I.~Evtimov and et~al., ``Robust physical-world attacks on deep learning
  models,'' \emph{arXiv:1707.08945}, 2017.

\bibitem{Distillation}
N.~Papernot and et~al., ``Distillation as a defense to adversarial
  perturbations against deep neural networks,'' in \emph{Security and Privacy
  (SP), 2016 IEEE Symposium on}, 2016, pp. 582--597.

\bibitem{Deepfool}
S.-M. Moosavi-Dezfooli and et~al., ``Deepfool: a simple and accurate method to
  fool deep neural networks,'' in \emph{Proceedings of the IEEE Conference on
  Computer Vision and Pattern Recognition}, 2016, pp. 2574--2582.

\bibitem{Robustness}
N.~Carlini and et~al., ``Towards evaluating the robustness of neural
  networks,'' in \emph{Security and Privacy (SP), 2017 IEEE Symposium on},
  2017, pp. 39--57.

\bibitem{MNIST}
Y.~LeCun, ``The mnist database of handwritten digits,'' \emph{http://yann.
  lecun. com/exdb/mnist/}, 1998.

\bibitem{cifar}
A.~Krizhevsky and et~al., ``The cifar-10 dataset,'' \emph{http://www. cs.
  toronto. edu/kriz/cifar. html}, 2014.

\bibitem{BP}
Y.~LeCun and et~al., ``Gradient-based learning applied to document
  recognition,'' \emph{Proceedings of the IEEE}, vol.~86, no.~11, pp.
  2278--2324, 1998.

\bibitem{Limitation}
N.~Papernot and et~al., ``The limitations of deep learning in adversarial
  settings,'' in \emph{Security and Privacy (EuroS\&P), 2016 IEEE European
  Symposium on}, 2016, pp. 372--387.

\bibitem{Tensorflow}
M.~Abadi and et~al., ``Tensorflow: Large-scale machine learning on
  heterogeneous distributed systems,'' \emph{arXiv:1603.04467}, 2016.

\bibitem{cleverhans}
P.~Nicolas and et~al., ``cleverhans v2.0.0: an adversarial machine learning
  library,'' \emph{arXiv:1610.00768}, 2017.

\end{thebibliography}
}
\end{document}